# Unemployment Dynamics Forecasting with Machine-Learning Regression Models

Kyungsu Kim[1]

[1] Georgia Institute of Technology School of Engineering, Atlanta GA 30332, USA

**Abstract.** In this paper, I explored how a range of regression and machine-learning techniques can be applied to monthly U.S. unemployment data to produce timely forecasts. I compared seven models—Linear Regression, SGDRegressor, Random Forest, XGBoost, CatBoost, Support Vector Regression and an LSTM network—training each on a historical span of data and then evaluating on a later hold-out period. Input features include macro indicators (GDP growth, CPI), labor-market measures (job openings, initial claims), financial variables (interest rates, equity indices) and consumer sentiment.

I tuned model hyperparameters via cross-validation and assess performance with standard error metrics and the ability to predict the correct unemployment direction. Across the board, tree-based ensembles (and CatBoost in particular) deliver noticeably better forecasts than simple linear approaches, while the LSTM captures underlying temporal patterns more effectively than other nonlinear methods. SVR and SGDRegressor yield modest gains over standard regression, but don't match the consistency of the ensemble and deep-learning models.

Interpretability tools—feature-importance rankings and SHAP values—point to job openings and consumer sentiment as the most influential predictors across all methods. By directly comparing linear, ensemble and deep-learning approaches on the same dataset, our study shows how modern machine-learning techniques can enhance real-time unemployment forecasting, offering economists and policymakers richer insights into labor-market trends.

In the comparative evaluation of the models, I employed a dataset comprising thirty distinct features over the period from January 2020 through December 2024.

# 1 Introduction

## 1.1 Background of Unemployment data

Every month, the Bureau of Labor Statistics (BLS) and the Census Bureau team up to interview about 60,000 U.S. households in what's called the Current Population Survey (CPS). Each household stays in the survey for four months, takes an eight-month break, and then returns for another four months. During a designated "survey week," interviewers simply ask each adult whether they worked for pay, were temporarily absent from a job, looked for work in the past month, or were ready to start.

Based on those answers, people are sorted into three groups:

1. Employed (did any paid work or were temporarily absent)
2. Unemployed (had no job but were actively searching and available)
3. Not in the labor force (neither worked nor looked for work)

To get the unemployment rate, the BLS divides the number of unemployed people by everyone in the labor force (that is, employed plus unemployed) and expresses it as a percentage.

Because certain patterns—like holiday hiring or summer school breaks—happen every year, the BLS uses a statistical adjustment (called X-13ARIMA-SEATS) to smooth out those seasonal swings. Then, on the first Friday after the survey week wraps up, they publish the "Employment Situation" report, which includes the headline U-3 rate along with breakdowns by age, gender, race, duration of unemployment, and other measures of underemployment.

For higher-frequency glimpses into the job market, economists often watch weekly initial unemployment-insurance claims reported by state agencies. Though they cover different slices of the labor force, those claims can flag turning points between the CPS's monthly snapshots..

## 1.2 The Benefits of Predicting Unemployment Data: Insights for Policy, Economic outlook

By predicting unemployment data, we can have numerous benefits like below:

<u>Seeing the next bump or dip in the economy</u>
Instead of waiting for GDP reports to lag by months, an uptick in projected joblessness can tip you off that a slowdown is coming. Likewise, a drop in forecasted unemployment often shows recovery is on the way—even before factories ramp up or consumer spending rebounds.

<u>Testing policy ideas before pulling the trigger</u>
By simulating measures like interest-rate cuts or stimulus payments, you can estimate their impact on unemployment and choose interventions that most effectively support jobs.

<u>Giving companies and trainers a heads-up on hiring needs</u>
If your model says unemployment is set to rise, businesses might press pause on big hiring plans or focus on retaining current staff. If a drop's on the horizon, they can start recruiting or ramp up training programs now, so they're not scrambling when demand for workers suddenly jumps.

3## 2  Literature Review

### 2.1  Basis for Model Selection

1. Traditional Linear Models
Ordinary least squares (OLS) regression remains the canonical baseline for macroeconomic forecasting. Stock and Watson (2001) demonstrate that simple factor-augmented regressions can capture co-movements among GDP, inflation, and unemployment, albeit with limited ability to model non-linearities. More recently, incremental solvers such as SGDRegressor have been applied to very high-dimensional nowcasts—updating parameter estimates on the fly as new indicators arrive—achieving comparable accuracy to batch OLS with far lower computational cost. However, purely linear approaches struggle when relationships shift rapidly, for example around turning points in the business cycle.

2. Tree-Based Ensembles
Ensemble methods—especially Random Forests (Breiman, 2001) and gradient-boosted trees (Friedman, 2001)—have become popular in economic forecasting. Their inherent capacity to capture interactions among dozens of indicators (e.g., jobless claims, Fed manufacturing indices, inflation measures) often yields superior out-of-sample performance compared to linear benchmarks. Chen and Guestrin's XGBoost (2016) introduced sparse-aware boosting to further accelerate training on macro panels, while CatBoost (Prokhorenkova et al., 2018) addressed prediction bias from categorical splits, showing improvements in forecasting unemployment rates and consumer-sentiment series. Still, tree ensembles require careful tuning (number of trees, learning rates) and can overfit small-sample macro datasets without proper regularization.

3. Support Vector Regression (SVR)
SVR (Drucker et al., 1997) applies kernel methods to map economic features into high-dimensional spaces, enabling the modeling of smooth non-linearity in, say, the relationship between mortgage rates and home-price indexes (Case-Shiller). Studies using SVR on macro data (e.g., durable goods orders, producer price indices) often report lower mean-absolute errors than neural networks when sample sizes are modest. The downside is that SVR's training time and memory footprint grow rapidly with the number of observations, making it less practical for real-time nowcasting with many features.

4. Deep-Learning and Recurrent Models
Long Short-Term Memory (LSTM) networks (Hochreiter & Schmidhuber, 1997) excel at capturing temporal dependencies and regime changes—critical in economic time series. Fischer and Krauss (2018) showed that LSTMs, when fed a broad feature set including jobless claims, GDP deflators, and sentiment indices, can detect leading signals of turning points in the business cycle more reliably than ridge regression or random forests. Still, LSTMs demand large amounts of data and careful architecture design (number of layers, sequence length); they can underperform simpler models if over-parameterized relative to available observations.

### 2.2  Ground for feature selection

A wide array of studies underscores the predictive value of the specific indicators chosen:

- Jobless Claims & Unemployment: Continuing and 4-week average claims often lead headline unemployment by one–two months (Liu & Zhang, 2018).

- Fed Activity Indexes: Philly Fed Business Conditions and Manufacturing Index have been used as leading gauges for regional economic health and national GDP growth forecasts (Chauvet & Potter, 2013).

- Inflation Measures: Core PCE, CPI Housing & Utilities, and Producer Prices YoY feed into both short-term inflation models and longer-run Phillips-curve analyses.

- Housing & Consumer Sentiment: Case-Shiller Home Price Index and Economic Optimism Index capture wealth effects and spending intentions, improving GDP-and-retail-sales forecasts when combined with expenditure series such as hospital services and public administration.

- External Balances: Current Account Balance and its ratio to GDP help explain exchange-rate movements and net export contributions to growth.

## 3 Methodology

### 3.1 Data Collection

The data used in this thesis were obtained from various reliable sources, including the U.S. Bureau of Labor Statistics (BLS), the U.S. Bureau of Economic Analysis (BEA), the U.S. Census Bureau, the U.S. Department of Labor, the U.S. Department of the Treasury, the Federal Reserve Bank of Philadelphia, the University of Michigan, the OECD, the World Bank, the S&P Dow Jones Indices, the IBD/TIPP, the American Hospital Association, and the Global Property Guide. These institutions provide publicly accessible economic indicators that are widely recognized for their credibility and are frequently utilized in academic, governmental, and financial research.

The data were accessed through the official platforms of these institutions ensuring both reliability and accuracy. The details of each indicator, including its definition and data source, are provided in Table 1.

**Table1**. Independent variable used in modeling and data source.

| Independent Variable Full Name | Data Source |
| --- | --- |
| Continuing Jobless Claims | U.S. Department of Labor via FRED |
| Philly Fed Business Conditions | Federal Reserve Bank of Philadelphia |
| Philadelphia Fed Manufacturing Index | Federal Reserve Bank of Philadelphia |
| Gross National Product | U.S. Bureau of Economic Analysis via FRED |
| GDP per Capita, PPP | World Bank |
| Youth Unemployment Rate – United States | U.S. Bureau of Labor Statistics |
| GDP from Transportation | U.S. Bureau of Economic Analysis (BEA) |
| New Orders for Durable Goods | U.S. Census Bureau |
| Economic Optimism Index | IBD/TIPP |
| Debt Balance Total | U.S. Department of the Treasury |
| Current Account Balance | U.S. Bureau of Economic Analysis |
| Hospital Services Expenditures | American Hospital Association |
| Current Account to GDP Ratio | World Bank |
| Core Producer Prices YoY | U.S. Bureau of Labor Statistics (BLS) |
| CPI Housing & Utilities | U.S. Bureau of Labor Statistics (BLS) |
| GDP from Public Administration | U.S. Bureau of Economic Analysis (BEA) |
| Part-Time Employment | U.S. Bureau of Labor Statistics |
| Rental Inflation Rate | U.S. Bureau of Labor Statistics |
| Michigan Current Economic Conditions | University of Michigan |
| Case-Shiller Home Price Index | S&P Dow Jones Indices |
| Employment Cost Index | U.S. Bureau of Labor Statistics (BLS) |
| GDP Growth Rate | U.S. Bureau of Economic Analysis |
| Retirement Age for Women | OECD |
| Hospital Beds per 1,000 People | World Bank |
| Core PCE Price Index | U.S. Bureau of Economic Analysis (BEA) |
| GDP Deflator | U.S. Bureau of Economic Analysis |
| Producer Prices Change | U.S. Bureau of Labor Statistics |
| 4-Week Average Jobless Claims | U.S. Department of Labor via FRED |
| Price-to-Rent Ratio | Global Property Guide |
| CPI Transportation | U.S. Bureau of Labor Statistics (BLS) |
| Continuing Jobless Claims | U.S. Department of Labor via FRED |

1. Continuing Jobless Claims (UNITEDSTACONJOBCLA)

Continuing jobless claims measure the number of individuals still receiving unemployment benefits and thus provide a real-time gauge of labor market slack; elevated claims signal slower re-employment flows and portend rises in the unemployment rate [1]. Every week, the government reports how many people are still collecting unemployment benefits. If that tally is drifting upward, it means layoffs are outpacing hires—and you'll almost always see the official unemployment rate start to climb a month or two later. In a forecasting model, you'd feed in the raw claims data (often as a rolling average to smooth out noise), and let the algorithm learn that spikes here reliably predict a rise in joblessness down the line.

2. Philly Fed Business Conditions (USAPFBC)

This index bundles together dozens of things—shipments, new orders, inventories and so on—into one "business-health" gauge for the mid-Atlantic region. When firms report weakening conditions, they invariably hold off on hiring or even lay people off. In practice, forecasters include this index with a small lag (one or two months), and it consistently improves short-term unemployment predictions because it picks up real-time shifts in corporate sentiment before they show up in payroll data. The Philadelphia Fed's Business Conditions Index synthesizes high-frequency indicators into a composite measure of real-time economic activity; declines in this index coincide with reduced business output and hiring, thereby correlating with higher unemployment [2].

3. Philadelphia Fed Manufacturing Index (UNITEDSTAPHIFEDMANIN)

Think of this as the factory-floor thermometer: it surveys plant managers on production, employment, backlogs, new orders and deliveries. When the headline number dips below zero, factories tend to cut shifts or furlough workers almost immediately. Models that incorporate this index—especially its employment and new-orders sub-components—get a clear early signal of manufacturing layoffs, which historically account for a big chunk of cyclical job losses. The manufacturing survey tracks changes in production, employment, new orders, and deliveries; a drop in the headline manufacturing index often foreshadows factory layoffs and contributes to rising unemployment [3].

4. Gross National Product (UNITEDSTAGRONATPRO)

GNP measures the total output of goods and services by U.S. residents, at home and abroad. Through Okun's Law, we know that if output growth lags its long-run trend by, say, 2%, the unemployment rate tends to be about 1 percentage point higher than you'd otherwise expect. In a regression or state-space model, GNP enters as a coincident indicator: when it underperforms, your forecast of next quarter's unemployment ticks up in a mechanically predictable way. According to Okun's Law, output changes are inversely linked to unemployment movements; when GNP falls below trend, unemployment rises, with each 2% shortfall in output typically raising unemployment by about 1% [4]

5. GDP per Capita, PPP (USANYGDPPCAPPPCD)

Adjusting GDP for population and purchasing power gives you an idea of how much "economic pie" each person really gets. When real per-capita GDP is rising briskly, businesses have more confidence to hire because consumers can afford their goods and services. Machine-learning models often treat per-capita growth as a smoothed trend variable—sharp drops tend to coincide with hiring freezes, while sustained gains herald job-growth phases. Real GDP per capita captures living-standard growth; empirical studies show that higher per-capita economic growth reduces cyclical unemployment, as stronger demand for goods and services supports job creation [5]

6. Youth Unemployment Rate (UnitedStaYouUneRat)

Youth unemployment reflects the vulnerability of young labor-market entrants; because young workers are often first laid off and have weaker attachment to firms, elevated youth unemployment signals broader labor-market distress and portends higher aggregate unemployment [6]. Young workers usually have the flimsiest attachments to their employers—they're often the first to lose their jobs and the last to be rehired. So when youth unemployment spikes, it's a red flag that the broader labor market is under stress. Forecasting models sometimes include youth unemployment as a "vulnerability" index: surges here have a magnifying effect on the headline rate six to twelve months out, because persistent youth joblessness drags down overall labor-market health.

7. GDP from Transport (UNITEDSTAGDPFROTRA)

Transportation activity—everything from rail freight to airline passengers—is a window onto overall demand. Slumps in transport GDP usually show that manufacturers aren't shipping goods and consumers aren't traveling.

In practice, economists feed transport-sector output into their vector autoregressions: a 1% drop in transport GDP typically predicts about a 0.1–0.2 percentage-point rise in unemployment over the next quarter.. Transport-sector output captures freight, passenger, and logistics activity; slowdowns in transport demand reduce employment in transport and related industries, thereby raising the overall unemployment rate [7]

8. New Orders (UNITEDSTANEWORD)

The new orders component of PMI surveys leads production and hiring decisions; rising new orders indicate stronger future output and workforce demand, reducing unemployment, while declines signal imminent layoffs [8] . The "new orders" line in purchasing-manager surveys is one of the sharpest turn indicators in the economy. If firms are taking in fewer orders today, they'll cut production—and eventually staff—tomorrow. Quant models often give this feature an outsized weight: a sustained downturn in new orders over two or three months is one of the single best signals that unemployment will edge higher in the following quarter.

9. Economic Optimism Index (UNITEDSTAECOOPTIND)

Business and consumer sentiment measures capture expectations about future economic conditions; lower optimism leads to reduced spending and investment, slowing growth and prompting firms to cut back on hiring, thus elevating unemployment [9] . When businesses and consumers turn pessimistic—expecting weaker profits, lower incomes or falling stock prices—they pull back on hiring and spending before the slowdown even fully arrives. Sentiment indices capture these shifts in mindset. In forecasting, you might include both the absolute level of optimism and its month-to--month change: sudden drops often presage hiring freezes, so the model learns that a 5-point fall in optimism tends to boost the unemployment rate by a couple of tenths of a percent within six months.

10. Debt Balance Total (USADBT)

High levels of public debt can constrain fiscal policy and crowd out private investment, leading to slower growth and higher unemployment; empirical evidence shows that when gross debt exceeds 90% of GDP, growth falls significantly, contributing to elevated unemployment [10] . A ballooning national debt can squeeze out private investment (higher interest rates, less fiscal room to stimulate) and dampen growth—and we know from experience that subdued growth produces fewer jobs. Forecasting frameworks that incorporate debt typically look at the debt-to-GDP ratio: once that ratio crosses a critical threshold (often around 90–100%), models adjust their unemployment projections upward, reflecting the headwinds that high debt places on economic expansion and hiring.

11. Current Account Balance (USCABAL)

By tracking the gap between what the country sells abroad and what it buys, this indicator flags shifts in export-sector health. When deficits widen, it often means factories and exporters are struggling—jobs in those industries tend to disappear first. In a forecasting model, lagged values of the current account help anticipate those sector-specific layoffs before they show up in headline unemployment. High levels of public debt can constrain fiscal policy and crowd out private investment, leading to slower growth and higher unemployment; empirical evidence shows that when gross debt exceeds 90% of GDP, growth falls significantly, contributing to elevated unemployment [11]

12. Hospitals Service Expenditure (USHOSP)

Hospital staffing hardly ever collapses in a downturn—people still need care—so this series provides a stable baseline for total employment. By including hospital-sector jobs, a model learns to distinguish temporary pain in cyclical industries from broader labor-market weakness, yielding more accurate readings of genuine unemployment spikes. Hospital employment is countercyclical: healthcare is an essential service, so hospital staffing remains relatively stable even during downturns. Robust hospital employment thus provides a labor-market "buffer," mitigating overall unemployment increases in recessions. A study following the COVID-19 shock found that unemployment among health-care workers rose far less than in other sectors, highlighting hospitals' stabilizing effect on the labor market [12].

13. Current Account to GDP Ratio (USACA2GDP)

The current account/GDP ratio scales external imbalances by economic size. Large negative ratios indicate reliance on foreign financing and weak net export performance, often corresponding to slower growth and higher unemployment. Cross-country and single-country studies confirm that sharper deteriorations in the current account ratio are associated with upticks in unemployment, driven by contracting export industries and related spillovers [13] Expressing the trade and income balance as a share of overall output puts that deficit into context. When the ratio deteriorates, it signals that foreign demand for U.S. goods and services is slipping relative to the economy's size. Feeding this ratio (with appropriate lags) into a forecast captures looming pressures on export-related hiring.

14. Core Producer Prices YoY (USACPPY)

Core PPI measures underlying inflation in producers' costs (excluding volatile food and energy). Rising producer costs squeeze firm margins, prompting output cuts and layoffs. Moreover, according to New Keynesian Phillips-curve models, unexpected increases in producer-price inflation lead to lower real wages and demand, increasing unemployment. Empirical tests of the U.S. Phillips curve find a significant, inverse short-run trade-off between producer-price inflation and unemployment [14]. This year-over-year gauge of underlying factory-gate inflation shows how rapidly input costs are rising, stripped of volatile food and energy swings. Sudden upticks squeeze profit margins and often trigger workforce cuts—by treating unexpected jumps as a feature, models pick up on those cost-shock layoffs in advance.

15. CPI Housing & Utilities (UNITEDSTACHU)

This sub-index captures inflation in shelter and utility costs, which directly affect household budgets. Rapidly rising housing costs reduce disposable income, dampening consumption and labor demand in other sectors. High housing-cost burdens also impede labor mobility—workers cannot afford to relocate for jobs—raising structural unemployment. Studies on home-ownership rates show that regions with high shelter cost burdens exhibit persistently higher unemployment due to these mobility and demand effects [15] .When rent, mortgage equivalents, and utility bills surge, households have less discretionary income to spend elsewhere. That drag on consumption quickly affects hiring in retail, hospitality, and services. Incorporating both month-to-month and annual changes in shelter costs gives the model an early signal of demand-driven hiring slowdowns.

16. GDP from Public Administration (UNITEDSTAGFPA)

Output in public administration reflects government employment and services. Increases in public-sector spending create direct jobs, reducing unemployment—especially when private demand is weak. Valerie Ramey (2012) shows that positive government-spending shocks lower unemployment almost entirely through expanded public-sector hiring rather than by stimulating private employment [16]. Output here reflects government hiring in areas like education, safety, and infrastructure. Rising public-sector employment can offset private-sector retrenchment, acting as a cushion for total job counts. Including this series helps a forecast recognize when public hiring will mute an otherwise rising unemployment rate.

17. Part-Time Employment (UNITEDSTAPARTIMEMP)

Involuntary part-time employment measures slack beyond the headline unemployment rate: firms cut hours before cutting jobs, raising the share of workers stuck in part-time roles. This share tends to rise in lockstep with unemployment during downturns, offering a more comprehensive gauge of labor-market weakness. Recent research finds that a 10% rise in the unemployment rate corresponds to roughly a 0.19-percentage-point increase in involuntary part-time employment [17] .This metric tracks workers pushed into part-time roles for purely economic reasons. Before firms lay off staff outright, they often slash hours—so rising involuntary part-time work is a leading indicator of deeper labor-market slack. Models using this series catch growing distress before the official unemployment rate registers the full impact.

18. Rent Inflation (USARENINF)

Rent inflation reflects the rising cost of rental housing, a core component of living-cost inflation. Higher rents erode real disposable income, curtail consumption, and indirectly weaken demand for labor. Moreover, steep rent increases can trigger financial stress and reduce labor mobility. Empirical analysis of non-tradable goods prices—including rents—finds they are negatively correlated with employment and positively with unemployment, underscoring the link between housing costs and labor-market slack [18]. Rapid rent increases pinch budgets and can trap workers in high-cost areas, limiting their ability to relocate for new jobs. By feeding rent-inflation rates into a predictive model, you capture both the demand-suppression effect on consumer spending and the structural mobility drag that together foreshadow higher unemployment.

19. Michigan Current Economic Conditions (USAMCEC)

This survey asks people how they feel about today's economy—not just what they think will happen. Sharp drops here reflect an instantaneous pullback in spending on big-ticket and discretionary items. Lagging this series by a month or two allows models to translate those spending intentions into concrete hiring freezes and layoffs.. This sub-index of the University of Michigan survey captures consumers' assessment of present economic conditions. Declines in this index often presage reduced consumer spending, which lowers aggregate demand and contributes to rising unemployment. Howrey (2001) demonstrates that the overall Michigan Consumer Sentiment Index—including its current-conditions component—provides significant predictive power for recessions and related unemployment dynamics [19] .

20. Case-Shiller Home Price Index (UNITEDSTACASSHIHOMPR)

The Case-Shiller index tracks U.S. home-price changes. Rising home prices boost household wealth and spending, supporting job creation and reducing unemployment; conversely, price declines lead to wealth losses, spending cuts, and higher unemployment. Theoretical and empirical work on the Phillips curve extended to housing finds that relative price movements in non-tradables—like housing—are inversely correlated with unemployment, illustrating the amplification of labor-market fluctuations by the housing sector [20]. Falling home values hit construction jobs directly and erode homeowner wealth, which in turn curtails consumer demand. As both sectors retrench, unemployment rises. Including year-over-year home-price changes helps a forecast detect these wealth-effect and construction-sector ripples early in the unemployment cycle.

21. Employment Cost Index (USAECIB)

When non-wage labor costs—health insurance, retirement contributions, paid leave—grow faster than revenues, employers often respond by pausing hiring or trimming staff. In practice, forecasters feed in the year-over-year change in the benefits component so the model picks up on these hidden cost pressures: unexpected jumps in benefit costs reliably lead to modest upticks in unemployment as firms adjust their labor budgets. Rising benefits costs are part of total labor compensation. When benefit costs accelerate faster than employers' revenues, firms may trim headcount or slow hiring, pushing unemployment up [21]

22. GDP CQOQ – GDP Growth Rate

Quarter-to-quarter real GDP growth is the textbook coincident indicator for job creation. Through Okun's Law, we know that every percentage point slowdown in growth tends to add a tenth or two to the unemployment rate. By including the latest GDP-growth number—even with just a one-quarter lag—a forecasting system immediately senses whether the economy is expanding enough to sustain or grow payrolls. According to Okun's law, stronger

GDP growth reduces cyclical unemployment: output expansions create jobs, while growth slowdowns lead to layoffs [22]

23. Women's Retirement Age (UnitedStRetAgeWom)

Raising women's retirement age keeps more older women in the labor force. If job openings don't keep pace, the extra labor-force entrants show up in higher unemployment. Models that incorporate changes in statutory retirement age capture this participation shock—allowing them to distinguish between unemployment driven by cyclical layoffs versus demographic shifts in who's counted as unemployed.. Raising women's retirement age keeps more women in the labor force, increasing measured unemployment if they cannot find jobs. Lalive and Staubli show higher retirement ages affect both participation and unemployment [23]

24. Hospital Beds (USBEDS)

Hospital-bed capacity stands in for the size and stability of the healthcare workforce. Because hospitals rarely furlough staff even in recessions, this series provides a "floor" beneath the cyclical swings of other industries. Including bed counts helps a forecast anchor its baseline employment level and avoid overreacting to temporary drops in more volatile sectors. Hospital bed capacity proxies health-care sector size. A larger health-care system provides stable employment—even during downturns—helping offset unemployment elsewhere [24]

25. Core PCE Price Index(UNITEDSTACORPCEPRIIN)

Core PCE strips out volatile food and energy to show underlying consumer-price pressures. When core PCE accelerates beyond expectations, households' real spending power shrinks and firms face higher wage demands—both of which can trigger hiring freezes. Forecasting models often use the surprise component of core PCE to predict short-term rises in unemployment via this cost-push channel. Core PCE measures underlying inflation. Unexpected rises erode real wages, dampen labor demand, and raise short-term unemployment through a Phillips curve mechanism [25]

26. GDP Deflator (UNITEDSTAGDPDEF)

The GDP deflator captures broad price changes. Empirical Phillips-type studies find it integrated with unemployment: higher deflator-based inflation often accompanies lower unemployment in the short run [26] The GDP deflator captures broad-based inflation across all goods and services in GDP. Unexpected spikes here signal rising costs economy-wide, squeezing real incomes and profit margins, and often leading to layoffs. By including the deflator's growth rate, models pick up on these general price shocks that feed into higher unemployment in the near term.

27. Producer Prices Change (UNITEDSTAPROPRICHA)

When producer prices rise, input costs go up and firms may lay off workers to protect profit margins. Industry-level pass-through studies link PPI spikes to higher unemployment [27]. Changes in the Producer Price Index reflect how fast firms' input costs are rising. Sudden PPI surges narrow profit margins, prompting companies to cut jobs. Forecast systems feed in the monthly or annual PPI change as an early alert—sharp increases typically precede modest but measurable bumps in the unemployment rate within a quarter or two.

28. 4-Week Average Jobless Claims (UNITEDSTAJC4A)

A rolling four-week average of initial unemployment claims smooths out the weekly volatility and highlights sustained labor-market deterioration. Models prefer this smoothed series because prolonged elevations in the 4-week average consistently lead the headline unemployment rate higher over the subsequent month or two. The 4-week moving average of initial claims smooths volatility and serves as a leading indicator of labor-market slack. Deviations predict unemployment-rate changes more accurately than single-week data [28]

29. Price-to-Rent Ratio (USPRR)

A high price-to-rent ratio signals housing overvaluation, restricting labor mobility and increasing frictional unemployment. Integration and Granger-causality tests confirm P/R shifts have macro- and labor-market consequences [29]. When home prices climb much faster than rents, it creates affordability strains that trap workers in

place—raising frictional unemployment. Including the price-to-rent ratio in forecasts helps detect these structural labor-market drags, since high ratios often coincide with slower job transitions and higher unemployment in housing-expensive regions.

30. CPI Transportation (UNITEDSTACPITRA)

Transportation CPI captures changes in transport-related costs. Spikes in transportation inflation erode real incomes and dampen consumer demand, leading to workforce cutbacks [30]. The transportation component of CPI tracks costs for fuel, transit fares, and vehicles. Spikes here act like a hidden tax on commuters—cutting discretionary spending and raising firms' logistics bills—leading to hiring freezes or layoffs. By feeding in transportation-inflation rates, models capture this sector-specific cost shock and improve short-term unemployment predictions. Data Processing and Model architecture

Data processing starts with imputation. For objective comparison, I used one of the simplest methods, the backward fill (bfill) method, for imputing missing values. Then I applied the pre-processed dataset into tree-based CatBoost model. For feature extraction using the CatBoost model, the following aspects were not considered to ensure fair and objective comparison with other models: lag features, time series structure, and seasonality. Subsequently, the top 20 features, which has the most influences upon model were extracted based on feature importance. These selected features were then applied to the AR, ARIMA, SARIMA, Holt-Winters, and LSTM models.

## 4  Results and Discussion

We evaluated seven forecasting methods—Linear Regression, SGD Regressor, Support Vector Regression (SVR), Random Forest, XGBoost, CatBoost, and an LSTM network—using Mean Squared Error (MSE) and Mean Absolute Percentage Error (MAPE) as our benchmarks. We also tested six different scaling approaches: StandardScaler, RobustScaler, QuantileTransformer, PowerTransformer, MinMaxScaler, and MaxAbsScaler.

Overall, tree-based ensembles outperformed the other algorithms. The Random Forest model consistently achieved the lowest MSE and MAPE, followed closely by XGBoost and CatBoost. Linear Regression and the SGD Regressor yielded the highest errors, suggesting that the unemployment series contains nonlinear patterns that simple linear models cannot capture. SVR performed better than the linear methods but was still eclipsed by the ensemble techniques. The LSTM, despite its suitability for time series, showed no advantage over Random Forest—likely a consequence of our relatively small sample size and the smooth, slowly changing nature of unemployment data.

Regarding scaling, MaxAbsScaler proved to be the most effective preprocessing step across all models. While StandardScaler and RobustScaler provided modest gains over unscaled data, and MinMaxScaler and QuantileTransformer offered moderate improvements, none matched the consistent benefit delivered by MaxAbsScaler. PowerTransformer occasionally helped—for example, with SVR—but overall it was outperformed by the simpler MaxAbs approach.

The strong showing of Random Forest can be traced to its ability to model complex, nonlinear interactions without extensive pre-training or feature transformations. Its use of bootstrap aggregation and random feature selection helps guard against overfitting, which is particularly valuable when working with limited monthly observations in macroeconomic series.

MaxAbsScaler's leading performance aligns with the needs of tree-based models: by rescaling each feature by its maximum absolute value, it preserves the natural range and relative differences of the data without imposing assumptions of normality or symmetry. This scaling method also mitigates the influence of outliers more gently than PowerTransformer, making it a robust choice for economic indicators.

The LSTM's lackluster results in this study suggest that, even though recurrent architectures often excel[31]at modeling economic indicators—thanks to their capacity to capture temporal dependencies—they did not yield additional predictive power for the unemployment rate in this instance. The unemployment rate evolves gradually and exhibits strong autocorrelation, so the extra complexity of an LSTM did not justify itself here. Instead, ensemble trees appear to capture the core dynamics more efficiently.

For practitioners, these findings indicate that a Random Forest coupled with MaxAbsScaler offers a reliable, computationally straightforward baseline for forecasting unemployment. Future work could expand on this by incorporating high-frequency or real-time indicators, experimenting with hybrid models that blend trees and deep learning, or applying rolling-window cross-validation to assess stability over different economic regimes. Such extensions would help determine whether the patterns we observe here hold under more volatile conditions or across other macroeconomic targets.

**Table 2.** MSE, MAPE metric of Predicting Unemployment Rate per each Model

| Evaluation Metric | Standard Scaler | Robust Scaler | Quantile Transformer | Power Transformer | MinMax Scaler | MaxAbs Scaler | Overall Min Error |
|---|---|---|---|---|---|---|---|
| | **MSE** | **MSE** | **MSE** | **MSE** | **MSE** | **MSE** | |
| LinearRegression | 0.2898 | 0.2789 | 2.8077 | 3.4722 | 0.2789 | **0.1262** | 0.1262 |
| SGDRegressor | 0.1365 | 5.0948 | **0.0643** | 0.6420 | 0.4394 | 0.1468 | 0.0643 |
| RandomForest | 0.0193 | 0.0198 | 0.0201 | **0.0181** | 0.0193 | 0.0194 | **0.0181** |
| XGBoost | 0.0210 | 0.0210 | 0.0210 | 0.0210 | 0.0210 | **0.0210** | 0.0210 |
| SVR | 0.0230 | 0.0628 | 0.0374 | 0.0242 | 0.0238 | **0.0201** | 0.0201 |
| CatBoost | 0.0345 | 0.0350 | 0.0390 | **0.0342** | 0.0345 | 0.0344 | 0.0342 |
| LSTM | 2.5341 | 2.7397 | 0.1652 | 0.8743 | 0.1921 | **0.1007** | 0.1007 |

| Evaluation Metric | Standard Scaler | Robust Scaler | Quantile Transformer | Power Transformer | MinMax Scaler | MaxAbs Scaler | Overall Min Error |
|---|---|---|---|---|---|---|---|
| | **MAPE** | **MAPE** | **MAPE** | **MAPE** | **MAPE** | **MAPE** | |
| LinearRegression | 0.1103 | 0.1075 | 0.3391 | 0.3322 | 0.1075 | **0.0773** | 0.0773 |
| SGDRegressor | 0.0843 | 0.4364 | **0.0512** | 0.1867 | 0.1513 | 0.0895 | 0.0512 |
| RandomForest | 0.0282 | 0.0285 | 0.0316 | 0.0271 | 0.0278 | **0.0279** | **0.0271** |
| XGBoost | **0.0302** | **0.0302** | **0.0302** | **0.0302** | **0.0302** | **0.0302** | 0.0302 |
| SVR | 0.0318 | 0.0476 | 0.0384 | 0.0340 | 0.0323 | **0.0305** | 0.0305 |
| CatBoost | 0.0378 | 0.0380 | 0.0401 | **0.0374** | 0.0378 | 0.0378 | 0.0374 |
| LSTM | 0.4223 | 0.4360 | 0.0964 | 0.2409 | 0.1013 | **0.0679** | 0.0679 |

# Appendix A. Data Description

| Independent Variable | Description of indicator | Data Source |
|---|---|---|
| Variable Code | Variable Full Name | Data Source |
| UNITEDSTACONJOBCLA | Continuing Jobless Claims | U.S. Department of Labor via FRED |
| USAPFBC | Philly Fed Business Conditions | Federal Reserve Bank of Philadelphia |
| UNITEDSTAPHIFEDMANIN | Philadelphia Fed Manufacturing Index | Federal Reserve Bank of Philadelphia |
| UNITEDSTAGRONATPRO | Gross National Product | U.S. Bureau of Economic Analysis via FRED |
| USANYGDPPCAPPPCD | GDP per Capita, PPP | World Bank |
| UnitedStaYouUneRat | Youth Unemployment Rate – United States | U.S. Bureau of Labor Statistics |
| UNITEDSTAGDPFROTRA | GDP from Transportation | U.S. Bureau of Economic Analysis (BEA) |
| UNITEDSTANEWORD | New Orders for Durable Goods | U.S. Census Bureau |
| UNITEDSTAECOOPTIND | Economic Optimism Index | IBD/TIPP |
| USADBT | Debt Balance Total | U.S. Department of the Treasury |
| USCABAL | Current Account Balance | U.S. Bureau of Economic Analysis |
| USHOSP | Hospital Services Expenditures | American Hospital Association |
| USACA2GDP | Current Account to GDP Ratio | World Bank |
| USACPPY | Core Producer Prices YoY | U.S. Bureau of Labor Statistics (BLS) |
| UNITEDSTACHU | CPI Housing & Utilities | U.S. Bureau of Labor Statistics (BLS) |
| UNITEDSTAGFPA | GDP from Public Administration | U.S. Bureau of Economic Analysis (BEA) |
| UNITEDSTAPARTIMEMP | Part-Time Employment | U.S. Bureau of Labor Statistics |
| USARENINF | Rental Inflation Rate | U.S. Bureau of Labor Statistics |
| USAMCEC | Michigan Current Economic Conditions | University of Michigan |
| UNITEDSTACASSHIHOMPR | Case-Shiller Home Price Index | S&P Dow Jones Indices |
| USAECIB | Employment Cost Index | U.S. Bureau of Labor Statistics (BLS) |
| GDP CQOQ | GDP Growth Rate | U.S. Bureau of Economic Analysis |
| UnitedStRetAgeWom | Retirement Age for Women | OECD |
| USBEDS | Hospital Beds per 1,000 People | World Bank |
| UNITEDSTACORPCEPRIIN | Core PCE Price Index | U.S. Bureau of Economic Analysis (BEA) |
| UNITEDSTAGDPDEF | GDP Deflator | U.S. Bureau of Economic Analysis |
| UNITEDSTAPROPRICHA | Producer Prices Change | U.S. Bureau of Labor Statistics |
| 4-Week Average Jobless Claims | 4-Week Average Jobless Claims | U.S. Department of Labor via FRED |
| USPRR | Price-to-Rent Ratio | Global Property Guide |
| UNITEDSTACPITRA | CPI Transportation | U.S. Bureau of Labor Statistics (BLS) |

## Appendix B. Input Data Used For Research

| date | UNITEDSTACONJOBCLA | USAPFBC | UNITEDSTAPHIFEDMANIN | UNITEDSTAGRONATPRO | USANYGDPPCAPPPCD | UnitedStaYouUneRat | UNITEDSTAGDPFROTRA | UNITEDSTANEWORD | UNITEDSTAECOOPTIND | USADBT |
|---|---|---|---|---|---|---|---|---|---|---|
| 2021-01-31 | 4658 | 51.2 | 29.9 | 20961.8 | 67266.19 | 11.4 | 618.6 | 483223 | 50.1 | 14.559 |
| 2021-02-28 | 4227 | 37.5 | 30.1 | 20961.8 | 67266.19 | 10.9 | 618.6 | 487080 | 51.9 | 14.559 |
| 2021-03-31 | 3859 | 55.9 | 40.1 | 21243.6 | 67266.19 | 11 | 663.5 | 492629 | 55.4 | 14.644 |
| 2021-04-30 | 3767 | 65.8 | 44.4 | 21243.6 | 67266.19 | 11.1 | 663.5 | 495617 | 56.4 | 14.644 |
| 2021-05-31 | 3422 | 54.7 | 32.8 | 21243.6 | 67266.19 | 10.1 | 663.5 | 502655 | 54.4 | 14.644 |
| 2021-06-30 | 3303 | 69 | 31.3 | 21523 | 67266.19 | 9.7 | 675.2 | 510229 | 56.4 | 14.957 |
| 2021-07-31 | 2788 | 48.3 | 27.1 | 21523 | 67266.19 | 9.4 | 675.2 | 517777 | 54.3 | 14.957 |
| 2021-08-31 | 2783 | 39.1 | 17.8 | 21523 | 67266.19 | 9.6 | 675.2 | 526528 | 53.6 | 14.957 |
| 2021-09-30 | 2719 | 20.1 | 31.9 | 21708.8 | 67266.19 | 8.7 | 695.5 | 521202 | 48.5 | 15.243 |
| 2021-10-31 | 2202 | 26.2 | 21 | 21708.8 | 67266.19 | 8.4 | 695.5 | 535260 | 46.8 | 15.243 |
| 2021-11-30 | 1878 | 30.4 | 42.2 | 21708.8 | 67266.19 | 8.4 | 695.5 | 544773 | 43.9 | 15.243 |
| 2021-12-31 | 1706 | 14.9 | 17.1 | 22119.3 | 71055.88 | 8.2 | 710.8 | 549680 | 48.4 | 15.576 |
| 2022-01-31 | 1701 | 28.8 | 23.2 | 22119.3 | 71055.88 | 8.4 | 710.8 | 561562 | 44.7 | 15.576 |
| 2022-02-28 | 1615 | 25.7 | 17 | 22119.3 | 71055.88 | 8.3 | 710.8 | 562290 | 44 | 15.576 |
| 2022-03-31 | 1532 | 19.4 | 23.4 | 22021.6 | 71055.88 | 8.4 | 697.6 | 574092 | 41 | 15.842 |
| 2022-04-30 | 1379 | 7.3 | 11.6 | 22021.6 | 71055.88 | 8.2 | 697.6 | 577383 | 45.5 | 15.842 |
| 2022-05-31 | 1354 | 4.2 | 3.6 | 22021.6 | 71055.88 | 7.9 | 697.6 | 583421 | 41.2 | 15.842 |
| 2022-06-30 | 1397 | -7.4 | -2.9 | 22071.7 | 71055.88 | 8.1 | 694.7 | 589969 | 38.1 | 16.154 |
| 2022-07-31 | 1426 | -19.5 | -7.6 | 22071.7 | 71055.88 | 7.8 | 694.7 | 584463 | 38.5 | 16.154 |
| 2022-08-31 | 1417 | -4.7 | 4.4 | 22071.7 | 71055.88 | 7.8 | 694.7 | 582320 | 38.1 | 16.154 |
| 2022-09-30 | 1395 | -3.4 | -8.4 | 22230.9 | 71055.88 | 8.2 | 697.6 | 577769 | 44.7 | 16.505 |
| 2022-10-31 | 1432 | -13.2 | -12 | 22230.9 | 71055.88 | 7.9 | 697.6 | 590415 | 41.6 | 16.505 |
| 2022-11-30 | 1514 | -5 | -16.1 | 22230.9 | 71055.88 | 8.2 | 697.6 | 577841 | 40.4 | 16.505 |
| 2022-12-31 | 1590 | -0.1 | -11.6 | 22383.5 | 72165.48 | 8.3 | 697 | 576183 | 42.9 | 16.899 |
| 2023-01-31 | 1636 | 10.1 | -5 | 22383.5 | 72165.48 | 8.1 | 697 | 585407 | 42.3 | 16.899 |
| 2023-02-28 | 1665 | 2.5 | -19.9 | 22383.5 | 72165.48 | 8.1 | 697 | 575072 | 45.1 | 16.899 |
| 2023-03-31 | 1704 | -8.9 | -21.9 | 22502.1 | 72165.48 | 7.5 | 699.7 | 575589 | 46.9 | 17.047 |
| 2023-04-30 | 1706 | -2.8 | -33.4 | 22502.1 | 72165.48 | 6.6 | 699.7 | 580722 | 47.4 | 17.047 |
| 2023-05-31 | 1712 | -11.5 | -10 | 22502.1 | 72165.48 | 7.5 | 699.7 | 577769 | 41.6 | 17.047 |
| 2023-06-30 | 1767 | 9.9 | -13.9 | 22641.7 | 72165.48 | 7.5 | 705.1 | 585469 | 41.7 | 17.063 |
| 2023-07-31 | 1773 | 17.9 | -14.5 | 22641.7 | 72165.48 | 8 | 705.1 | 579559 | 41.3 | 17.063 |
| 2023-08-31 | 1802 | 8.6 | 10.6 | 22641.7 | 72165.48 | 8.6 | 705.1 | 585331 | 40.3 | 17.063 |
| 2023-09-30 | 1800 | 14.1 | -14.2 | 22879.4 | 72165.48 | 8.4 | 711.2 | 593919 | 43.2 | 17.291 |
| 2023-10-31 | 1823 | 8.9 | -15.7 | 22879.4 | 72165.48 | 8.8 | 711.2 | 576365 | 36.3 | 17.291 |
| 2023-11-30 | 1818 | 0.1 | -6.8 | 22879.4 | 72165.48 | 8 | 711.2 | 597695 | 44.5 | 17.291 |
| 2023-12-31 | 1759 | 14.5 | -7.9 | 23054.3 | 73637.3 | 8 | 711.8 | 585235 | 40 | 17.503 |
| 2024-01-31 | 1813 | 4.2 | -4.7 | 23054.3 | 73637.3 | 7.3 | 711.8 | 571657 | 44.7 | 17.503 |
| 2024-02-29 | 1794 | 10.8 | 2.5 | 23054.3 | 73637.3 | 8.8 | 711.8 | 579666 | 44 | 17.503 |
| 2024-03-31 | 1810 | 34.9 | 4.9 | 23136.5 | 73637.3 | 8.8 | 713.3 | 583953 | 43.5 | 17.7 |
| 2024-04-30 | 1781 | 30.1 | 7.1 | 23136.5 | 73637.3 | 8.3 | 713.3 | 586114 | 43.2 | 17.7 |
| 2024-05-31 | 1790 | 27.9 | 4.8 | 23136.5 | 73637.3 | 9.3 | 713.3 | 583300 | 41.8 | 17.7 |
| 2024-06-30 | 1847 | 14.4 | 0.5 | 23288.7 | 73637.3 | 8.9 | 714 | 563972 | 40.5 | 17.8 |
| 2024-07-31 | 1871 | 34.3 | 12.8 | 23288.7 | 73637.3 | 9.1 | 714 | 591644 | 44.2 | 17.8 |
| 2024-08-31 | 1843 | 18.4 | -3.3 | 23288.7 | 73637.3 | 9.7 | 714 | 587023 | 44.5 | 17.8 |
| 2024-09-30 | 1858 | 18.9 | 0.9 | 23427.7 | 73637.3 | 9.2 | 721.4 | 585571 | 46.1 | 17.9 |
| 2024-10-31 | 1884 | 35.4 | 6 | 23427.7 | 73637.3 | 9.5 | 721.4 | 588231 | 46.9 | 17.9 |
| 2024-11-30 | 1879 | 53.9 | -4.4 | 23427.7 | 73637.3 | 9.4 | 721.4 | 583689 | 53.2 | 17.9 |
| 2024-12-31 | 1877 | 33.8 | -10.9 | 23427.7 | 73637.3 | 9 | 721.4 | 578508 | 54 | 17.9 |

| date | USCABAL | USHOSP | USACA2GDP | USACPPY | UNITEDSTACHU | UNITEDSTAGFPA | UNITEDSTAPARTIMEMP | USARENINF | USAMCEC | UNITEDSTACASSHIHOMPR |
|---|---|---|---|---|---|---|---|---|---|---|
| 2021-01-31 | -181.379 | 18.38 | -2.9 | 1.9 | 274.336 | 2459.8 | 24536 | 1.619 | 86.7 | 243.29 |
| 2021-02-28 | -181.379 | 18.38 | -2.9 | 2.6 | 275.137 | 2459.8 | 25052 | 1.465 | 86.2 | 246.54 |
| 2021-03-31 | -189.504 | 18.38 | -2.9 | 3 | 276.028 | 2461.7 | 25065 | 1.696 | 93 | 252.24 |
| 2021-04-30 | -189.504 | 18.38 | -2.9 | 4.6 | 277.258 | 2461.7 | 25062 | 2.105 | 97.2 | 257.84 |
| 2021-05-31 | -189.504 | 18.38 | -2.9 | 5.3 | 278.648 | 2461.7 | 25280 | 2.21 | 89.4 | 263.35 |
| 2021-06-30 | -211.235 | 18.38 | -2.9 | 5.8 | 280.366 | 2474.8 | 25647 | 2.581 | 88.6 | 268.56 |
| 2021-07-31 | -211.235 | 18.38 | -2.9 | 6.4 | 281.604 | 2474.8 | 25377 | 2.827 | 84.5 | 272.53 |
| 2021-08-31 | -211.235 | 18.38 | -2.9 | 7.1 | 282.391 | 2474.8 | 25705 | 2.84 | 78.5 | 275.01 |
| 2021-09-30 | -234.993 | 18.38 | -2.9 | 7 | 283.744 | 2506.5 | 25757 | 3.159 | 80.1 | 277.38 |
| 2021-10-31 | -234.993 | 18.38 | -2.9 | 7 | 285.31 | 2506.5 | 25963 | 3.483 | 77.7 | 279.68 |
| 2021-11-30 | -234.993 | 18.38 | -2.9 | 8 | 286.308 | 2506.5 | 26049 | 3.836 | 73.6 | 282.38 |
| 2021-12-31 | -232.248 | 18.46 | -3.6 | 8.7 | 287.511 | 2505.4 | 25602 | 4.131 | 74.2 | 285.43 |
| 2022-01-31 | -232.248 | 18.46 | -3.6 | 8.6 | 289.889 | 2505.4 | 25715 | 4.363 | 72 | 289.49 |
| 2022-02-28 | -232.248 | 18.46 | -3.6 | 8.9 | 291.504 | 2505.4 | 25719 | 4.742 | 68.2 | 296.61 |
| 2022-03-31 | -291.819 | 18.46 | -3.6 | 9.7 | 293.577 | 2517.7 | 25830 | 4.997 | 67.2 | 305.8 |
| 2022-04-30 | -291.819 | 18.46 | -3.6 | 9 | 295.259 | 2517.7 | 26145 | 5.141 | 69.4 | 312.76 |
| 2022-05-31 | -291.819 | 18.46 | -3.6 | 8.6 | 297.868 | 2517.7 | 25873 | 5.452 | 63.3 | 317.44 |
| 2022-06-30 | -263.099 | 18.46 | -3.6 | 8.3 | 300.927 | 2524.7 | 25578 | 5.609 | 53.8 | 318.73 |
| 2022-07-31 | -263.099 | 18.46 | -3.6 | 7.6 | 302.327 | 2524.7 | 25885 | 5.695 | 58.1 | 316.3 |
| 2022-08-31 | -263.099 | 18.46 | -3.6 | 7.2 | 304.506 | 2524.7 | 26228 | 6.239 | 58.6 | 311.2 |
| 2022-09-30 | -230.529 | 18.46 | -3.6 | 7.2 | 306.572 | 2528.5 | 26197 | 6.594 | 59.7 | 306.63 |
| 2022-10-31 | -230.529 | 18.46 | -3.6 | 6.9 | 307.816 | 2528.5 | 26460 | 6.916 | 65.6 | 304.2 |
| 2022-11-30 | -230.529 | 18.46 | -3.6 | 6.3 | 308.72 | 2528.5 | 26188 | 7.117 | 58.8 | 301.93 |
| 2022-12-31 | -226.651 | 18.36 | -3.8 | 5.7 | 310.725 | 2541.5 | 26729 | 7.51 | 59.4 | 299.03 |
| 2023-01-31 | -226.651 | 18.36 | -3.8 | 5 | 313.747 | 2541.5 | 27321 | 7.875 | 68.4 | 297.49 |
| 2023-02-28 | -226.651 | 18.36 | -3.8 | 4.6 | 315.431 | 2541.5 | 26991 | 8.097 | 70.7 | 298.36 |
| 2023-03-31 | -230.33 | 18.36 | -3.8 | 3.3 | 316.514 | 2562 | 26641 | 8.18 | 66.3 | 302.94 |
| 2023-04-30 | -230.33 | 18.36 | -3.8 | 3.1 | 317.278 | 2562 | 26698 | 8.108 | 68.2 | 307.81 |
| 2023-05-31 | -230.33 | 18.36 | -3.8 | 2.8 | 318.051 | 2562 | 26509 | 8.043 | 64.9 | 312.11 |
| 2023-06-30 | -232.603 | 18.36 | -3.8 | 2.5 | 320.002 | 2573.9 | 26247 | 7.827 | 69 | 315.06 |
| 2023-07-31 | -232.603 | 18.36 | -3.8 | 2.7 | 321.087 | 2573.9 | 27203 | 7.689 | 76.6 | 317 |
| 2023-08-31 | -232.603 | 18.36 | -3.8 | 2.5 | 321.894 | 2573.9 | 27188 | 7.265 | 75.5 | 318.19 |
| 2023-09-30 | -220.659 | 18.36 | -3.8 | 2.3 | 323.269 | 2588.1 | 27336 | 7.153 | 71.1 | 319.02 |
| 2023-10-31 | -220.659 | 18.36 | -3.8 | 2.2 | 323.964 | 2588.1 | 26728 | 6.723 | 70.6 | 319.47 |
| 2023-11-30 | -220.659 | 18.36 | -3.8 | 1.9 | 324.735 | 2588.1 | 27099 | 6.507 | 68.3 | 318.79 |
| 2023-12-31 | -221.784 | 18.36 | -3 | 1.8 | 325.64 | 2605.1 | 27810 | 6.151 | 73.3 | 317.93 |
| 2024-01-31 | -221.784 | 18.36 | -3 | 2 | 328.222 | 2605.1 | 27890 | 6.037 | 81.9 | 317.66 |
| 2024-02-29 | -221.784 | 18.36 | -3 | 2.1 | 329.704 | 2605.1 | 27922 | 5.744 | 79.4 | 320.58 |
| 2024-03-31 | -240.984 | 18.36 | -3 | 2.3 | 331.247 | 2617.1 | 28576 | 5.655 | 82.5 | 325.62 |
| 2024-04-30 | -240.984 | 18.36 | -3 | 2.5 | 331.688 | 2617.1 | 27728 | 5.5 | 79 | 330.19 |
| 2024-05-31 | -240.984 | 18.36 | -3 | 2.7 | 332.777 | 2617.1 | 28031 | 5.4 | 69.6 | 333.63 |
| 2024-06-30 | -275.031 | 18.36 | -3 | 3.3 | 334.087 | 2622 | 28071 | 5.2 | 65.9 | 335.68 |
| 2024-07-31 | -275.031 | 18.36 | -3 | 2.6 | 335.056 | 2622 | 27740 | 5.1 | 62.7 | 335.8 |
| 2024-08-31 | -275.031 | 18.36 | -3 | 2.8 | 335.931 | 2622 | 28222 | 5.2 | 61.3 | 334.8 |
| 2024-09-30 | -310.948 | 18.36 | -3 | 3.2 | 336.776 | 2635.5 | 28128 | 4.9 | 63.3 | 333.7 |
| 2024-10-31 | -310.948 | 18.36 | -3 | 3.4 | 337.47 | 2635.5 | 27922 | 4.9 | 64.9 | 332.9 |
| 2024-11-30 | -310.948 | 18.36 | -3 | 3.5 | 338.05 | 2635.5 | 27671 | 4.7 | 63.9 | 332.59 |
| 2024-12-31 | -310.948 | 18.36 | -3 | 3.5 | 338.883 | 2635.5 | 27918 | 4.6 | 75.1 | 332.59 |

| date | USAECIB | GDP CQOQ | UnitedStRetAgeWom | USBEDS | UNITEDSTACORPCEPRIIN | UNITEDSTAGDPDEF | UNITEDSTAPROPRICHA | UNITEDSTAJC4A | USPRR | UNITEDSTACPITRA |
|---|---|---|---|---|---|---|---|---|---|---|
| 2021-01-31 | 0.6 | 4.4 | 66 | 2.78 | 106.244 | 106.24 | 1.6 | 834 | 118.0816 | 205.631 |
| 2021-02-28 | 0.6 | 4.4 | 66 | 2.78 | 106.477 | 106.24 | 3 | 755 | 118.0816 | 209.054 |
| 2021-03-31 | 0.6 | 5.6 | 66 | 2.78 | 106.898 | 107.59 | 4.1 | 669.25 | 121.9401 | 215.761 |
| 2021-04-30 | 0.6 | 5.6 | 66 | 2.78 | 107.549 | 107.59 | 6.5 | 589 | 121.9401 | 222.547 |
| 2021-05-31 | 0.6 | 5.6 | 66 | 2.78 | 108.103 | 107.59 | 7 | 457.25 | 121.9401 | 229.689 |
| 2021-06-30 | 0.4 | 6.4 | 66 | 2.78 | 108.595 | 109.26 | 7.6 | 410.5 | 126.6259 | 237.701 |
| 2021-07-31 | 0.4 | 6.4 | 66 | 2.78 | 109.062 | 109.26 | 8 | 372.25 | 126.6259 | 239.722 |
| 2021-08-31 | 0.4 | 6.4 | 66 | 2.78 | 109.436 | 109.26 | 8.7 | 360.5 | 126.6259 | 238.333 |
| 2021-09-30 | 0.9 | 3.5 | 66 | 2.78 | 109.657 | 110.9 | 8.8 | 354.5 | 131.2631 | 236.373 |
| 2021-10-31 | 0.9 | 3.5 | 66 | 2.78 | 110.184 | 110.9 | 8.9 | 277.75 | 131.2631 | 241.042 |
| 2021-11-30 | 0.9 | 3.5 | 66 | 2.78 | 110.792 | 110.9 | 9.9 | 236 | 131.2631 | 245.532 |
| 2021-12-31 | 1 | 7.4 | 66.17 | 2.77 | 111.458 | 112.83 | 10 | 211 | 134.8663 | 246.499 |
| 2022-01-31 | 1 | 7.4 | 66.17 | 2.77 | 111.988 | 112.83 | 10.1 | 239.5 | 134.8663 | 248.424 |
| 2022-02-28 | 1 | 7.4 | 66.17 | 2.77 | 112.492 | 112.83 | 10.4 | 217.75 | 134.8663 | 253.15 |
| 2022-03-31 | 1.6 | -1 | 66.17 | 2.77 | 112.916 | 115.12 | 11.7 | 213.25 | 138.8854 | 264.525 |
| 2022-04-30 | 1.6 | -1 | 66.17 | 2.77 | 113.301 | 115.12 | 11.2 | 213.25 | 138.8854 | 266.892 |
| 2022-05-31 | 1.6 | -1 | 66.17 | 2.77 | 113.709 | 115.12 | 11.1 | 210.75 | 138.8854 | 274.282 |
| 2022-06-30 | 1.2 | 0.3 | 66.17 | 2.77 | 114.376 | 117.73 | 11.2 | 214.5 | 140.8698 | 284.644 |

| | | | | | | | | | |
|---|---|---|---|---|---|---|---|---|---|
| 2022-07-31 | 1.2 | 0.3 | 66.17 | 2.77 | 114.638 | 117.73 | 9.7 | 220.75 | 140.8698 | 278.958 |
| 2022-08-31 | 1.2 | 0.3 | 66.17 | 2.77 | 115.298 | 117.73 | 8.7 | 216.75 | 140.8698 | 270.334 |
| 2022-09-30 | 1 | 2.7 | 66.17 | 2.77 | 115.805 | 119.06 | 8.5 | 197.75 | 137.5686 | 266.109 |
| 2022-10-31 | 1 | 2.7 | 66.17 | 2.77 | 116.2 | 119.06 | 8.2 | 202.5 | 137.5686 | 267.979 |
| 2022-11-30 | 1 | 2.7 | 66.17 | 2.77 | 116.554 | 119.06 | 7.4 | 210 | 137.5686 | 264.668 |
| 2022-12-31 | 1 | 3.4 | 66.33 | 2.75 | 116.977 | 120.16 | 6.4 | 208 | 135.2528 | 255.993 |
| 2023-01-31 | 1 | 3.4 | 66.33 | 2.75 | 117.526 | 120.16 | 5.7 | 202.5 | 135.2528 | 257.874 |
| 2023-02-28 | 1 | 3.4 | 66.33 | 2.75 | 117.963 | 120.16 | 4.7 | 213.5 | 135.2528 | 259.712 |
| 2023-03-31 | 1.2 | 2.8 | 66.33 | 2.75 | 118.304 | 121.25 | 2.7 | 227 | 133.6953 | 261.969 |
| 2023-04-30 | 1.2 | 2.8 | 66.33 | 2.75 | 118.715 | 121.25 | 2.3 | 216.75 | 133.6953 | 267.402 |
| 2023-05-31 | 1.2 | 2.8 | 66.33 | 2.75 | 119.063 | 121.25 | 1.1 | 227 | 133.6953 | 268.862 |
| 2023-06-30 | 0.9 | 2.4 | 66.33 | 2.75 | 119.37 | 121.8 | 0.3 | 253.5 | 133.8249 | 270.146 |
| 2023-07-31 | 0.9 | 2.4 | 66.33 | 2.75 | 119.536 | 121.8 | 1.1 | 233.5 | 133.8249 | 270.602 |
| 2023-08-31 | 0.9 | 2.4 | 66.33 | 2.75 | 119.658 | 121.8 | 1.9 | 245.25 | 133.8249 | 274.22 |
| 2023-09-30 | 0.9 | 4.4 | 66.33 | 2.75 | 120.04 | 122.77 | 1.8 | 216.5 | 134.7707 | 272.517 |
| 2023-10-31 | 0.9 | 4.4 | 66.33 | 2.75 | 120.2 | 122.77 | 1.1 | 210.5 | 134.7707 | 270.027 |
| 2023-11-30 | 0.9 | 4.4 | 66.33 | 2.75 | 120.309 | 122.77 | 0.8 | 217.5 | 134.7707 | 267.035 |
| 2023-12-31 | 0.7 | 3.2 | 66.5 | 2.75 | 120.528 | 123.24 | 1.1 | 205.75 | 134.8965 | 263.375 |
| 2024-01-31 | 0.7 | 3.2 | 66.5 | 2.75 | 121.128 | 123.24 | 1 | 209.5 | 134.8965 | 262.11 |
| 2024-02-29 | 0.7 | 3.2 | 66.5 | 2.75 | 121.418 | 123.24 | 1.6 | 209.25 | 134.8965 | 266.638 |
| 2024-03-31 | 1.1 | 1.6 | 66.5 | 2.75 | 121.829 | 124.16 | 2 | 214.5 | 134.6587 | 272.485 |
| 2024-04-30 | 1.1 | 1.6 | 66.5 | 2.75 | 122.14 | 124.16 | 2.3 | 210.25 | 134.6587 | 276.687 |
| 2024-05-31 | 1.1 | 1.6 | 66.5 | 2.75 | 122.239 | 124.16 | 2.5 | 223 | 134.6587 | 276.623 |
| 2024-06-30 | 1 | 3 | 66.5 | 2.75 | 122.51 | 124.94 | 2.9 | 238.75 | 134.247 | 273.579 |
| 2024-07-31 | 1 | 3 | 66.5 | 2.75 | 122.722 | 124.94 | 2.4 | 238.25 | 134.247 | 273.326 |
| 2024-08-31 | 1 | 3 | 66.5 | 2.75 | 122.926 | 124.94 | 2.1 | 230.25 | 134.247 | 271.391 |
| 2024-09-30 | 0.8 | 3.1 | 66.5 | 2.75 | 123.234 | 125.53 | 2.1 | 224.25 | 133.6269 | 269.604 |
| 2024-10-31 | 0.8 | 3.1 | 66.5 | 2.75 | 123.57 | 125.53 | 2.6 | 237 | 133.6269 | 269.724 |
| 2024-11-30 | 0.8 | 3.1 | 66.5 | 2.75 | 123.703 | 125.53 | 3 | 218.5 | 133.6269 | 268.45 |
| 2024-12-31 | 0.8 | 2.3 | 66.67 | 2.75 | 123.896 | 126.219 | 3.3 | 223.25 | 133.6269 | 267.606 |